\documentclass[10pt,twocolumn,letterpaper]{article}

\usepackage{iccv}              % To produce the CAMERA-READY version
\definecolor{iccvblue}{rgb}{0.21,0.49,0.74}
\usepackage[pagebackref,breaklinks,colorlinks,allcolors=iccvblue]{hyperref}

\def\paperID{8317} % *** Enter the Paper ID here
\def\confName{ICCV}
\def\confYear{2025}

\title{Multi-View Slot Attention Using Paraphrased Texts for Face Anti-Spoofing}
\author{Jeongmin Yu$^{1}$\thanks{Equal contribution} , Susang Kim$^{1, 2}$\footnotemark[1] , Kisu Lee$^1$, Taekyoung Kwon$^1$, Won-Yong Shin$^1$, Ha Young Kim$^1$\thanks{Corresponding author}\\
$^1$Yonsei University \quad
$^2$POSCO DX\\
{\tt\small \{jeongminyu, healess, kisu0928, taekyoung, wy.shin, hayoung.kim\}@yonsei.ac.kr}}
\usepackage{makecell}
\usepackage{algorithm}   
\usepackage[noend]{algpseudocode} % hide end for operation
\usepackage{multirow}
\usepackage{listings} % 코드 시퀀스 (appendix)
\usepackage{xcolor} % 코드 하이라이트를 위해 필요 (appendix)
\usepackage{xr} % Cross-reference 패키지

\begin{document}
\maketitle
\begin{abstract}
Recent face anti-spoofing (FAS) methods have shown remarkable cross-domain performance by employing vision-language models like CLIP.
However, existing CLIP-based FAS models do not fully exploit CLIP’s patch embedding tokens, failing to detect critical spoofing clues.
Moreover, these models rely on a single text prompt per class (\eg, `live' or `fake'), which limits generalization.
To address these issues, we propose MVP-FAS, a novel framework incorporating two key modules: Multi-View Slot attention (MVS) and Multi-Text Patch Alignment (MTPA).
Both modules utilize multiple paraphrased texts to generate generalized features and reduce dependence on domain-specific text.
MVS extracts local detailed spatial features and global context from patch embeddings by leveraging diverse texts with multiple perspectives.
MTPA aligns patches with multiple text representations to improve semantic robustness.
Extensive experiments demonstrate that MVP-FAS achieves superior generalization performance, outperforming previous state-of-the-art methods on cross-domain datasets.
Code: \href{https://github.com/Elune001/MVP-FAS}{https://github.com/Elune001/MVP-FAS}.
\end{abstract}    
\begin{figure}[ht]
  \centering 
  \includegraphics[width=\linewidth]{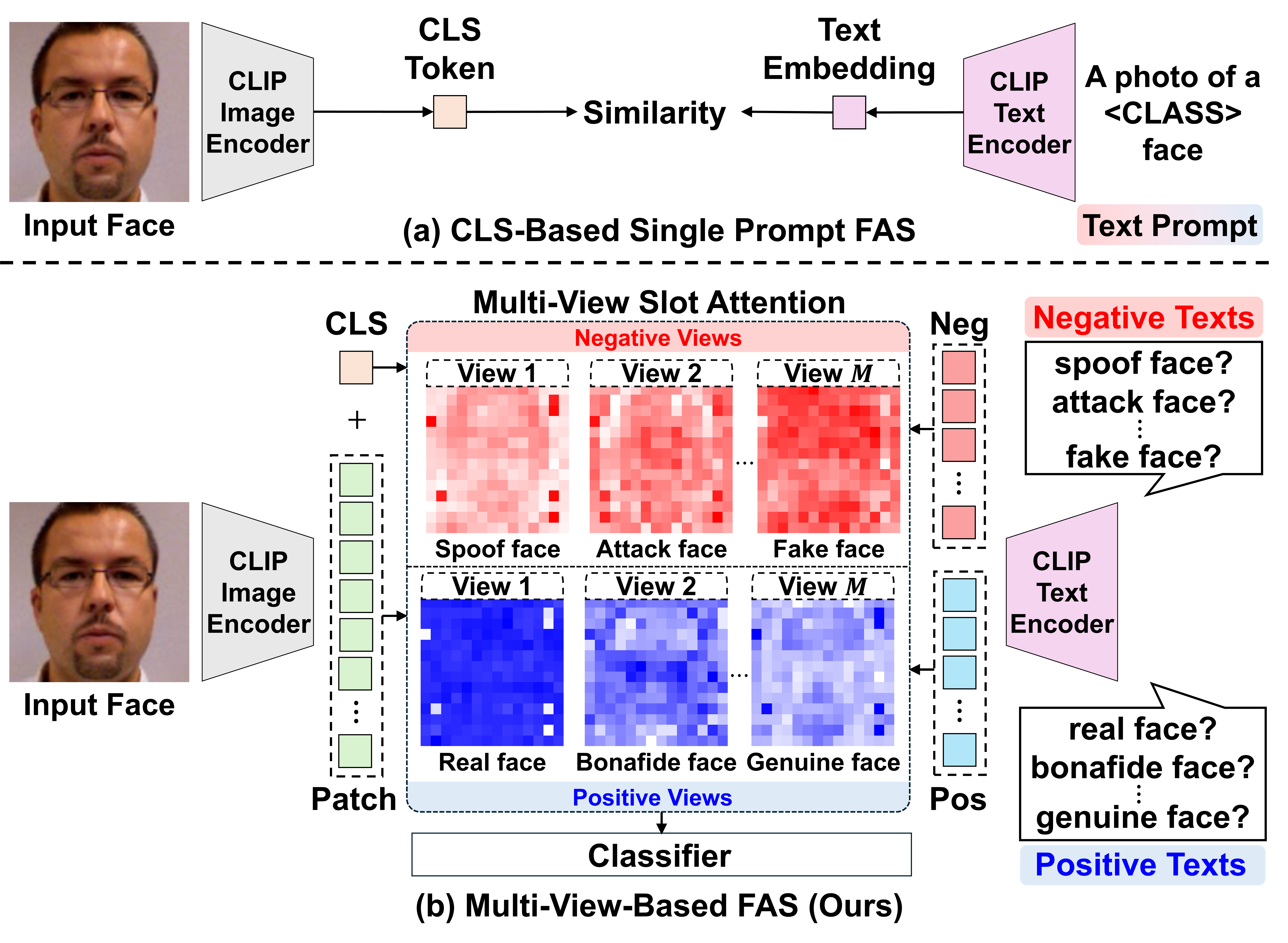} 
  \vspace{-0.6cm}
  \caption{
  \textbf{Comparison between CLS-based single prompt and multi-view-based FAS methods.} 
    (a) illustrates the single prompt FAS model using the CLIP CLS token, which captures global information but lacks fine-grained local details. 
    (b) represents our multi-view-based FAS framework (MVP-FAS), providing local details with global context for each view. The red and blue areas indicate where the model focuses on within each view.
  }
  \label{fig:MVSlotAttentionConecpt}
  \vspace{-0.6cm}
\end{figure}
\section{Introduction}
\label{sec:intro}
Face anti-spoofing (FAS), which distinguishes spoofed faces from real ones, has become essential for enhancing the security of facial recognition systems. 
Previous FAS models~\cite{li2016generalized,boulkenafet2015face,de2013lbp,komulainen2013context,yang2013face,patel2016secure,boulkenafet2016face,li2016original} have demonstrated notable effectiveness against various attack types, such as printed image~\cite{zhang2012face}, replay~\cite{chingovska2012effectiveness}, and 3D mask attacks~\cite{greenberg2017hackers} in intra-domain datasets.
However, in real-world scenarios, FAS models may encounter unseen attack types and data not present during training~\cite{shao2019multi,jia2020single,wang2020unsupervised}.
Therefore, extracting generalized features is crucial to enhancing FAS models' ability to accurately detect spoofed faces across a broader range of conditions~\cite{hu2024rethinking,huang2024one,lin2024suppress,zhou2024test,le2024gradient,liu2024cfpl}.

Recently, FAS methods~\cite{srivatsan2023flip,liu2024cfpl,liubottom,hu2024fine,guo2024style, chen2025mixture, li2025fa, fang2024unified} that leverage vision-language models (VLMs), such as CLIP~\cite{radford2021learning}, have demonstrated remarkable performance in zero-shot learning and domain generalization across diverse cross-domain attacks. 
The first CLIP-based FAS model, FLIP~\cite{srivatsan2023flip}, primarily relies on the classification token (CLS token), which encodes global image information.
Recent CLIP-based models~\cite{hu2024fine,liu2024cfpl,guo2024style} have proposed methods to utilize local spoofing clues.
FGPL~\cite{hu2024fine} captures local details through prompt learning and a convolutional adapter applied to the CLIP image encoder.
CFPL~\cite{liu2024cfpl} and S-CPTL~\cite{guo2024style} leverage spatial information by extracting style- or content-related features from patch embeddings and feeding them into the text encoder.
However, these methods~\cite{hu2024fine,liu2024cfpl,guo2024style} still do not directly utilize CLIP's patch embedding tokens, which contain rich local information within the image domain.
Consequently, similar to FLIP~\cite{srivatsan2023flip}, these approaches struggle to capture fine-grained spoofing artifacts, such as abnormal textures and light reflections in small areas, due to the information loss when projecting into the text embedding space for prompt generation. 
In addition, CLIP is trained to align the CLS token with the text embedding without applying direct supervision to the patch embeddings during training.
Thus, explicitly aligning local patch embeddings with text embeddings is necessary for effectively leveraging fine-grained local information.

Furthermore, as shown in previous studies~\cite{liu2024cfpl,hu2024fine,guo2024style,liubottom}, the input text prompt is another important factor in CLIP-based FAS.
These works have explored prompt learning methods that integrate image features, enhancing text descriptions beyond fixed prompts.
However, they still use only a single fixed text pair to represent the real and spoof classes, such as `live' and `fake'.
Since vanilla CLIP is not specifically trained for the FAS task, there is no guarantee that the class texts used to represent ‘spoof’ or ‘real’ have semantics that are generally applicable in FAS.
For instance, the word `live' can indicate a real face in FAS but can also mean live stream.
Therefore, as shown in prior research~\cite{srivatsan2023flip}  and our observation (see Sec.~\ref{sec:prompt_analysis}), relying on a single fixed text prompt per class can limit the model's robustness, affecting generalization performance.

To address these challenges, we propose a novel framework, \textbf{M}ulti-\textbf{V}iew cli\textbf{P} for \textbf{FAS} (\textbf{MVP-FAS}), which comprises two main modules: Multi-View Slot attention (\textbf{MVS}) and Multi-Text Patch Alignment (\textbf{MTPA}).
Both modules leverage various paraphrases of the class text to enhance generalization.
MVS extracts multi-view representations that capture local spoofing details and global context based on slot attention, where global-aware CLIP patch embeddings serve as queries and paraphrased text embeddings as keys and values.
As shown in Fig.~\ref{fig:MVSlotAttentionConecpt}~(b), MVS enables the model to interpret patches from various perspectives of multiple texts, leading to more generalized features.
MTPA aligns patch embeddings with a multi-text anchor derived from paraphrased texts, which improves alignment robustness by alleviating the impact of biased text representations in the embedding space.
By incorporating soft-masking, MTPA effectively aligns patches that are relatively important for spoofing prediction.

We evaluate MVP-FAS through extensive experiments on two protocols: Protocol 1~\cite{wen2015face, zhang2012face, chingovska2012effectiveness, boulkenafet2017oulu} and Protocol 2~\cite{zhang2020casia,liu2021casia,george2019biometric}, both designed for cross-domain datasets.
MVP-FAS achieves state-of-the-art (SOTA) cross-domain generalization performance by a large margin on both protocols in FAS.
Notably, TPR@FPR=1\% is consistently high overall, demonstrating its reliability in high-security scenarios.
Furthermore, our method visualizes attention scores for each class, supporting better decision-making and highlighting its potential for enhanced interpretability.
To summarize, our main contributions are as follows:
\begin{itemize}
    \item We introduce %design 
    MVS to extract generalized multi-view features by using multi-text with slot attention, enabling them to capture detailed local and global information from patch embeddings across diverse perspectives.
    \item We design %introduce 
    MTPA, which aligns CLIP’s patch embeddings with multiple paraphrased texts using soft-masking, improving semantic robustness and alignment effectiveness.
    \item Our model achieves SOTA performance across various cross-domain FAS datasets, demonstrating that integrating MVS and MTPA enhances generalization.
    \item We visualize the multi-view attention weights from MVS, offering more informative, region-based insights about the predictions than existing visualization methods.
\end{itemize}
\begin{figure*}[t]
  \centering 
  \includegraphics[width=0.88\linewidth]
  {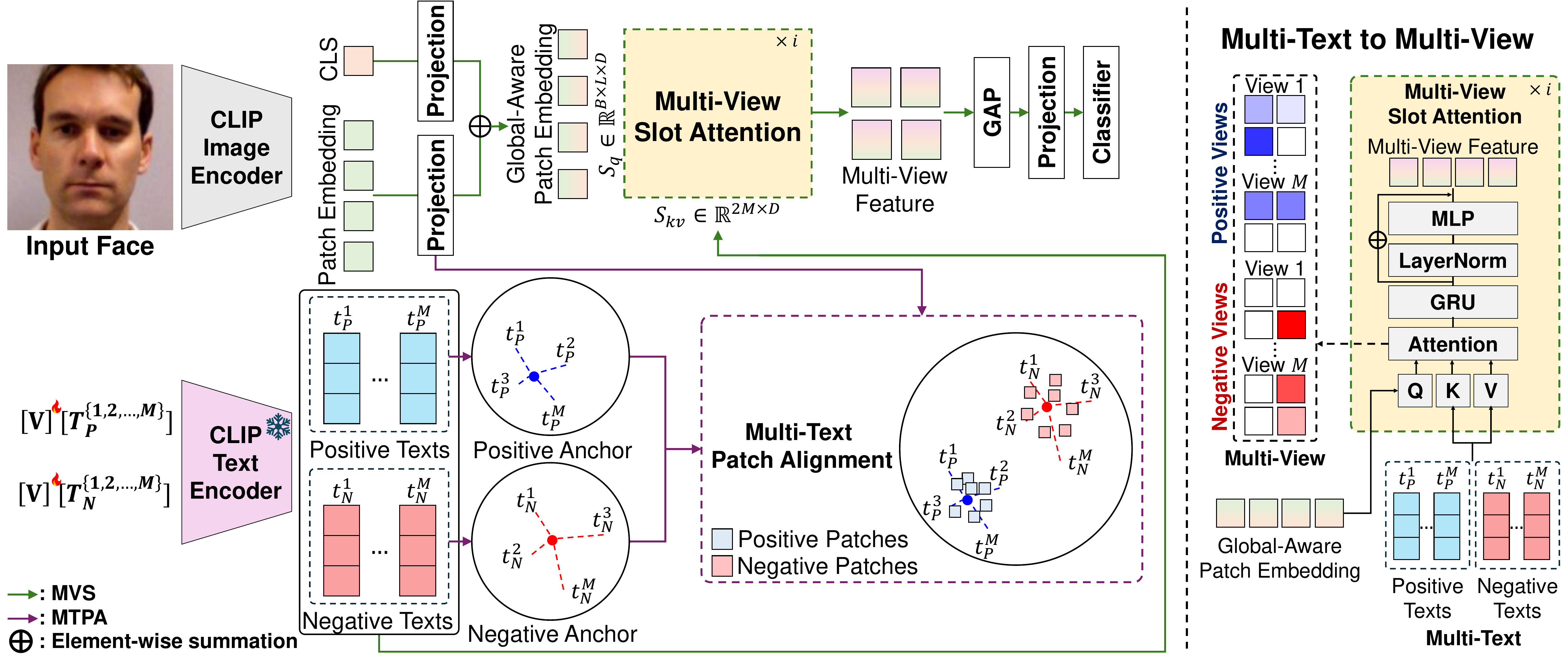}
\caption{\textbf{The overall architecture of MVP-FAS. }$t$ indicates the text embedding of the input prompt after passing through the text encoder.
  }
  \vspace*{-.4cm}
  \label{fig:OverallArchitecture}
\end{figure*}
\section{Related Work}
\label{sec:related work}

\noindent\textbf{Face Anti-Spoofing. }FAS has been studied for over a decade driven by increasingly sophisticated spoofing techniques. 
Early methods~\cite{boulkenafet2015face,de2013lbp,komulainen2013context,yang2013face,patel2016secure,boulkenafet2016face} relied on handcrafted features to identify counterfeits~\cite{pan2007eyeblink,li2016generalized}. With advances in deep learning, FAS has evolved into an end-to-end learning using convolutional neural networks for feature extraction and classification~\cite{li2016original,yu2020searching}. 
However, these approaches often struggle with cross-domain generalization.
To address this, domain adaptation techniques~\cite{li2018unsupervised,liu2022source} attempt to minimize distributional differences between the source and target domains, though their effectiveness depends on the availability of unlabeled target data. An alternative is domain generalization~\cite{wang2022domain,sun2023rethinking,le2024gradient,liu2021adaptive,liu2021dual,zhou2023instance}, which leverages multiple source domains without requiring target data. However, most existing methods rely on labeled images, limiting their ability to generalize across diverse real-world conditions.

\noindent\textbf{CLIP-based FAS. }CLIP~\cite{radford2021learning} has driven significant progress in VLMs, inspiring advancements across various domains such as visual question answering~\cite{tan2024compound}, semantic segmentation~\cite{zhou2023zegclip,mukhoti2023open}, and domain generalization~\cite{vidit2023clip}.
In FAS, various CLIP-based models~\cite{liu2024cfpl,guo2024style,hu2024fine,liubottom,srivatsan2023flip} have been proposed with prompt learning, employing CLIP's zero-shot transfer learning capabilities.
FLIP~\cite{srivatsan2023flip} is the first to utilize heuristic positive and negative prompt pairs for fine-tuning CLIP models, aligning real and fake images with corresponding prompts to enhance overall FAS performance.
CFPL~\cite{liu2024cfpl} introduces class-free prompt learning, emphasizing content and style of the face.
FGPL~\cite{hu2024fine} employs domain-agnostic information through prompt learning, while S-CPTL~\cite{guo2024style} reflects visual style information via prompt learning.
BUDoPT~\cite{liubottom} addresses varying domain prompt levels, including recording settings and attack types.
However, these methods rely on CLIP's patch embeddings within the prompt learning framework, potentially causing information loss compared to direct utilization. Consequently, they may struggle to effectively capture local spoofing clues in the image domain.

\noindent\textbf{Object-Centric Learning. }Object-centric learning has received substantial interest for its ability to generate semantically disentangled representations.
The slot attention mechanism~\cite{locatello2020object} groups the features with similar characteristics around a slot query in an object-centric manner, allowing each slot to represent a distinct object part.
Several studies~\cite{xu2023learning,fan2023unsupervised,huo2023geovln,yang2021self,sajjadi2022object,xu2025slot} have employed image tokens within the slot attention to cluster similar features effectively. Slot-VLM~\cite{xu2025slot} effectively bridges the gap between vision and language by leveraging object-centric and event-centric slots.
GeoVLN~\cite{huo2023geovln} extends slot-based visual representation learning to enhance robust navigation, while the binding module~\cite{xu2023learning} uses slot attention to cluster image tokens into groups. 
Inspired by these approaches, we first introduce a variant of slot attention with a multi-text approach to enhance generalization in FAS.

\noindent\textbf{Patch-Level Alignment in CLIP. }
In CLIP, the visual encoder's CLS token and the text encoder's features are well aligned for image-text pairs.
However, CLIP mainly learns global information~\cite{radford2021learning} during training, making it struggle to encode fine-grained details in patch embeddings.
Recent studies propose methods for aligning CLIP’s patch embeddings with text to enhance detailed information~\cite{mukhoti2023open,Bica2024ImprovingFU,liu2024fm}.
PACL~\cite{mukhoti2023open} introduces a modified CLIP's contrastive loss to enhance the alignment between visual patch embeddings and text encoder features, while FM-CLIP~\cite{liu2024fm} attempts to guide patch-level representations by adding text features into the patches, weighted according to their similarity.
Ke Fan \textit{et al.}~\cite{fan2023unsupervised} also show that fine-tuning CLIP with cross-attention using CLS token and patch embeddings enhances object localization.
While previous studies focused on single patch-text alignment, aligning multiple semantically equivalent texts with patches has been unexplored, despite its potential to enhance robustness.
\section{Method}
\subsection{Overall Architecture}
The architecture of MVP-FAS is shown in Fig.~\ref{fig:OverallArchitecture}.
MVP-FAS leverages the CLIP image encoder $\mathcal{V}(\cdot)$ and text encoder $\mathcal{T}(\cdot)$, where the text encoder is frozen to retain pre-trained linguistic knowledge, while the image encoder is fine-tuned to adapt to the FAS task.

\noindent\textbf{Text Encoder. }To obtain multi-view features, we utilize positive texts~$T_P=\{T^{j}_{P}\}^{M}_{j=1}$ for real faces and negative texts~$T_N=\{T^{j}_{N}\}^{M}_{j=1}$ for spoof faces, where $M$ is the number of views. 
All positive and negative texts are generated using ChatGPT~\cite{achiam2023gpt} as paraphrases of the texts `real face' and `spoof face'. 
The input prompt~$\mathcal{P}$ for the text encoder consists of a class text $\text{[CLASS]} \in T_P \cup T_N$ and context texts $[\mathbf{V}]$, as shown in Eq.~\eqref{eqn: learnable_prompt}.
Specifically, we employ the learnable prompt of CoOp~\cite{zhou2022learning} for the context texts to find more suitable text representations for FAS.
\begin{equation}
\label{eqn: learnable_prompt}
{\mathcal{P}} = [\mathbf{V}][\text{CLASS}].
\end{equation}
As a result, the set of input prompts are fed into the text encoder, resulting in the output multi-text embeddings $S_{kv} \in \mathbb{R}^{2M\times D}$, where~$D$ denotes the embedding dimension.

\noindent\textbf{Image Encoder. }Image encoder receives a batch of face images~$I\in\mathbb{R}^{B \times H \times W \times 3} $ as input and extracts a CLS token and patch embeddings, where $B$ is the batch size and $H,W$ are the image resolutions.
They are each passed through separate projection layers, each consisting of two-layer multilayer perceptron (MLP) to adjust dimensions to $D$.
Subsequently, the CLS token is broadcast across all patch embeddings, creating global-aware patch embeddings~$S_{q} \in \mathbb{R}^{B \times L\times D}$, where $L$ denotes the patch embedding length.

\noindent\textbf{Feature Extraction and Classification. }Finally, MVP-FAS leverages MVS to transform the global-aware patch embedding into generalized multi-view features, integrating global context and local detailed information based on multi-text embeddings from the text encoder.
The resulting multi-view features are then fed into a classification head, which comprises global average pooling, a two-layer MLP as a projection layer, and a fully connected layer as a classifier to predict whether the face is real or spoofed.

\begin{algorithm}[t]
\caption{Multi-View Slot Attention Module. }
\begin{algorithmic}[1] 
    \State \textbf{Input:} $S_{q}\in \mathbb{R}^{B \times L \times D}$, ${S}_{kv} \in \mathbb{R}^{2M \times D}$
        
    \State ${S}_{kv} = \text{repeat}({S}_{kv}, \text{dim=0}, \text{repeats=}B) \in \mathbb{R}^{B \times 2M \times D}$
    \State ${S}_{kv} = \text{LayerNorm}({S}_{kv})$
    \State $k, v = \text{Linear}_{k}({S}_{kv}),  \text{Linear}_{v}({S}_{kv})$
    
    \For{$i = 1$ to $i_{max}$}
        \State $S_{q_{\text{prev}}} = S_{q}$
        \State $q = \text{Linear}_{q}(\text{LayerNorm}(S_{q}))$
        \State $attn = \text{softmax}(q \cdot k^T \frac{1}{\sqrt{D}}, \text{axis=`query'})$
        \\ \textcolor{gray}{\hfill{\# \; Multi-Text to Multi-View}}
        \State $attn = attn / (\sum(attn, \text{axis=`key'}) + \epsilon)$
        \State $updates_{MV} = attn \cdot v$\textcolor{gray}{\qquad \;\;\# \hfill{Aggregate Views}}
        \State $S_{q} = \text{GRU}(updates_{MV}, S_{q_{\text{prev}}})$\textcolor{gray}{\;\# \hfill{Feature Update}}
        \State $S_{q} = S_{q} + \text{MLP}(\text{LayerNorm}(S_{q}))$ 
    \EndFor
    \State $\text{F}_{MV} = S_{q}$%\textcolor{gray}{\hfill{\# Multi-View Feature}}
    \State \Return $\text{F}_{MV}$ \Comment{Return Multi-View Feature}
\end{algorithmic}
\label{algorithm:MV_slot_attention}
\end{algorithm}

\subsection{Multi-View Slot Attention for VLMs}
Previous works~\cite{ liu2024cfpl,guo2024style} project image-level detailed information into the text embedding space via prompts, which can lead to the loss of visual characteristics.
To address this problem, MVS directly utilizes CLIP's patch embeddings based on slot attention~\cite{locatello2020object} to preserve the fine-grained details of local patches.

Unlike conventional slot attention, our MVS uses global-aware patch embeddings as queries and text embeddings as keys and values.
This design enables the global assignment of text embeddings while preserving the details of local patches, preventing information loss when patches are assigned to learnable queries in the standard slot attention.
As described in Algorithm~\ref{algorithm:MV_slot_attention}, to allocate text embedding to each global-aware patch embedding, MVS projects global-aware patch embeddings into patch queries~$q\in\mathbb{R}^{B\times L\times D}$ and text embeddings into text keys~$k\in\mathbb{R}^{B\times 2M\times D}$ and values~$v\in\mathbb{R}^{B\times 2M\times D}$. 
Then, MVS calculates the similarity between queries and keys through a scaled dot product and applies softmax along the query dimension~$L$ to compute the attention scores~$attn_{(q,k)}\in\mathbb{R}^{B\times L\times 2M}$.
\begin{equation}
\label{eqn: MVS_attn_score}
attn_{(q,k)} = \text{softmax}(q\cdot k^{\text{T}}\frac{1}{\sqrt{D}},\text{axis=`query'}).
\end{equation}
These attention scores are used to assign positive and negative class texts to the patches.

\noindent\textbf{Single-view ($M=1$).}
\begin{figure}[t]
  \centering 
  \includegraphics[width=\linewidth]
  {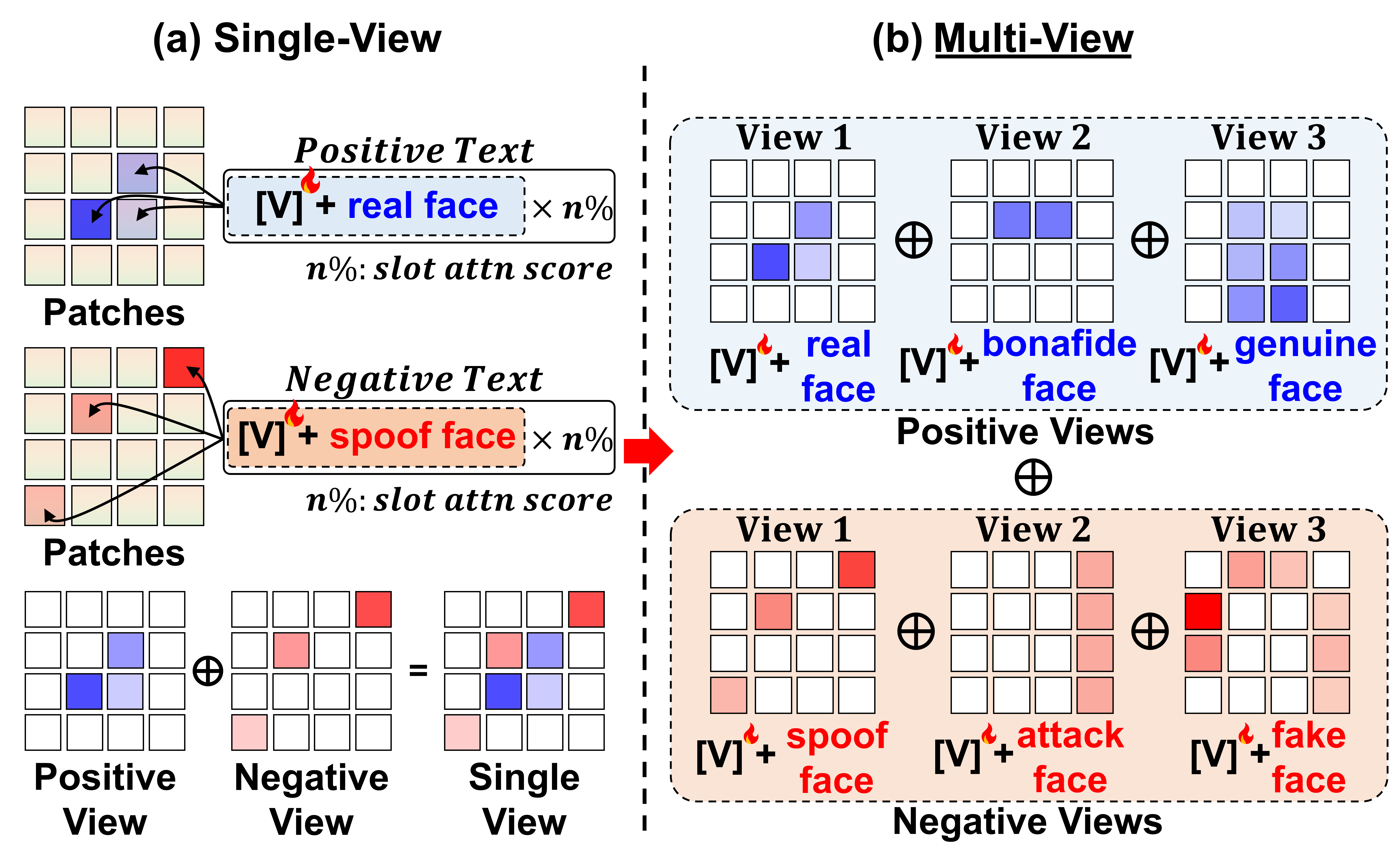}
  \vspace{-0.6cm}
    \caption{\textbf{The mechanism for binding text embeddings to patch embeddings in Multi-View Slot attention (MVS). }
        (a) depicts a single-view setting with $M=1$, while (b) illustrates a multi-view example where $M=3$.
    Each text embedding is distributed across patches based on the slot attention score of $n\%$. 
    All positive-negative view pairs are aggregated via element-wise summation ($\oplus$), resulting in the final multi-view output.
    }
  \vspace{-0.4cm}
  \label{fig:MultiViewSlotAttention}
\end{figure}
Guided by attention scores, a single pair of positive and negative text values is allocated across all patch queries.
As shown in Fig.~\ref{fig:MultiViewSlotAttention}~(a), global-aware patches~$S_q$ are updated with the update value~$\Delta S_q$, which is separated text values~$attn_{(q,k)}v$ and serves as input to the GRU~\cite{cho2014learning}, while $S_q$ is used as its hidden state.
\begin{equation}
\label{eqn: text_assign}
{S_q}' = \text{GRU}(\Delta S_q, S_q), \quad \text{where  } \Delta S_q= attn_{(q,k)}v.
\end{equation}
Specifically, the updated patches ${S_q}'$ are refined through the residual MLP in Algorithm~\ref{algorithm:MV_slot_attention}, expressed as follows:
\begin{equation}
\label{eqn: text_assign}
{S_q}' \leftarrow{S_q}' + \text{MLP(LayerNorm(${S_q}'$))}.
\end{equation}
This slot attention process preserves local patch information, which is crucial for spoofing prediction, employing a GRU when updating additional text values. 
By leveraging the semantics of distributed text values, the model effectively captures the global context among local patches while preserving essential patch information.
Concretely, as shown at the bottom of Fig.~\ref{fig:MultiViewSlotAttention}~(a), the positive and negative views derived from a single text pair are combined to form a single-view representation, which considers an image from both perspectives. %
\begin{equation}
\label{eqn: single_view}
\Delta S_q = attn_{(q,k_{p})}v_{p} + attn_{(q,k_{n})}v_{n},
\end{equation}
where ($k_{p}$, $k_{n}$) and ($v_{p}$, $v_{n}$) are keys and their corresponding values of positive-negative text embeddings, respectively. 
Through this process, the model focuses on semantically meaningful regions in the single-view representation.
However, a limitation of single-view slot attention is that the model's performance depends on the choice of prompt.

\noindent\textbf{Multi-view ($M>1$). }To improve generalization, MVS leverages paraphrased texts with the same meaning, similar to rephrasing search keywords to retrieve more useful information online. 
As illustrated in Fig.~\ref{fig:MultiViewSlotAttention}~(b), MVS takes multiple positive and negative text pairs as input and generates various positive and negative views corresponding to multi-text pairs. 
By combining these views, MVS generates multi-view representations that enable the model to capture more generalized and robust features.

Specifically, the update value of global-aware patch embeddings~$\Delta S_q$ can be represented as a weighted summation based on attention scores corresponding to the positive and negative text values $v\in \mathbb{R}^{B\times 2M\times D}$.

\begin{equation}
\label{eqn: MVS_patch_level_explanation}
\Delta S_q = \sum_{j=1}^{2M}attn_{(q,k_{j})}v_{j},
\end{equation}
where $v_{j}$ represents the $j$-th text embedding values and $attn_{(q,k_{j})}$ denotes the attention score between the patch queries and the $j$-th text embedding values.

\subsection{Multi-Text Patch Alignment for MVS}
Patch-level local information contains various detailed cues for FAS~\cite{wang2022patchnet}. 
While some CLIP-based methods~\cite{liu2024cfpl,hu2024fine,guo2024style} leverage patch embeddings to capture these details, the lack of patch-level alignment limits the effective utilization of local information. 

To address this, we introduce MTPA to ensure effective patch alignment while mitigating the impact of biased text semantics.
MTPA applies soft masking to patches with low similarity to the multi-text anchor using soft masking scores within the range (0, 1).
These soft-masked patches are processed by an auxiliary classifier, which enhances alignment by providing supervision that increases the similarity between informative patches relevant to spoofing prediction and their corresponding anchors.

Specifically, as shown in Fig.~\ref{fig:OverallArchitecture}, to determine the anchors of the patch alignment, we calculate the mean value of the CLIP text embeddings for the real and spoof classes.
\begin{equation}
\label{eqn: positive_and_negative_mean}
\begin{split}
A_{P} &= \frac{1}{M}\sum_{j=1}^{M}\mathcal{T}([\mathbf{V}][T^{j}_{P}]), \\A_{N} &= \frac{1}{M}\sum_{j=1}^{M}\mathcal{T}([\mathbf{V}][T^{j}_{N}]),
\end{split}
\end{equation}
where $A_{P} \in \mathbb{R}^{D}$ and $A_{N} \in \mathbb{R}^{D}$ represent the positive and negative anchors, respectively.

Subsequently, the cosine similarity $C\in \mathbb{R}^{B \times L \times 1}$ between the patch embedding~$E \in \mathbb{R}^{B \times L \times D}$ and the alignment anchor $A \in \{A_P, A_N\}^B$, which is selected based on the class label of each patch, is calculated as follows:

\begin{equation}
\label{eqn: similarity_patch_anchor}
C=E_{norm} \cdot A_{norm},
\end{equation}
where $E_{norm} \in \mathbb{R}^{B \times L \times D}$ is the normalized $E$, and $A_{norm} \in \mathbb{R}^{B \times D \times 1}$ is normalized and expanded $A$.

Finally, we apply a sigmoid~$\sigma(\cdot)$ to the cosine similarity and use the output as soft masking scores~$\mathcal{M}$ to suppress patches with lower similarity.
\begin{equation}
\label{eqn: softmask}
\mathcal{M} =\sigma(\alpha  C), 
\end{equation}
where $\alpha$ is a scaling factor of the cosine similarity, controlling the range of $\mathcal{M}$. In this paper, we set $\alpha$ to 10 to ensure that $\mathcal{M}$ spans the range (0,1).

These soft-masked patches are aggregated by summation and passed through a fully connected layer FC with softmax~$\mathcal{S}$ to obtain the spoofing probability~$p$, which is expressed as in the following:
\begin{equation}
\label{eqn: patch_emphasizing}
p =\mathcal{S} \left( \text{FC}\left(\sum_{l=1}^{L}\left(\mathcal{M}\times E_{l}\right)\right) \right). 
\end{equation}
Then, cross-entropy (CE) loss with the ground-truth~$y$ is computed for MTPA, as formulated in Eq.~\eqref{eqn: MTPA_final}.

\begin{equation}
\label{eqn: MTPA_final}
\mathcal{L}_{\text{MTPA}}=\frac{1}{B}\sum_{k=1}^{B}CE(y_{k}, p_{k}).
\end{equation}

\begin{table*}[ht!]
	\centering
    \setlength\tabcolsep{2.5pt}\resizebox{\linewidth}{!}{
\begin{tabular}{lccccccccccccc}
\toprule
\multirow{3}{*}{Method} & \multicolumn{3}{c}{OCI $\rightarrow$ M} & \multicolumn{3}{c}{OMI $\rightarrow$ C} & \multicolumn{3}{c}{OCM $\rightarrow$ I} & \multicolumn{3}{c}{ICM $\rightarrow$ O} & {avg.} \\ 
\cmidrule{2-14} & HTER $\downarrow$ & AUC $\uparrow$ & \makecell{TPR@\\FPR=1\%} $\uparrow$ & HTER $\downarrow$ & AUC $\uparrow$ & \makecell{TPR@\\FPR=1\%} $\uparrow$ & HTER $\downarrow$ & AUC $\uparrow$ & \makecell{TPR@\\FPR=1\%} $\uparrow$ & HTER $\downarrow$ & AUC $\uparrow$ & \makecell{TPR@\\FPR=1\%} $\uparrow$ & HTER $\downarrow$ \\ 
\midrule
MADDG (CVPR'19)~\cite{shao2019multi} & 17.69 & 88.06 & - & 24.50 & 84.51 & - & 22.19 & 84.99 & - & 27.98 & 80.02 & - & 23.09 \\
DR-MD-Net (TFIS'20)~\cite{wang2020unsupervised} & 17.02 & 90.10 & - & 19.68 & 87.43 & - & 20.87 & 86.72 & - & 25.02 & 81.47 & - & 20.64 \\
RFMeta (AAAI'20)~\cite{shao2020regularized} & 13.89 & 93.98 & - & 20.27 & 88.16 & - & 17.30 & 90.48 & - & 16.45 & 91.16 & - & 16.97 \\
NAS-FAS (TPAMI'20)~\cite{yu2020fas} & 19.53 & 89.36 & - & 16.54 & 90.11 & - & 14.51 & 93.84 & - & 13.80 & 93.47 & - & 16.09 \\
D$^2$AM (AAAI'21)~\cite{chen2021generalizable} & 12.70 & 95.66 & - & 20.98 & 85.58 & - & 15.43 & 91.22 & - & 15.27 & 90.87 & - & 16.09 \\
SDA (AAAI'21)~\cite{wang2021self} & 15.40 & 91.80 & - & 24.50 & 84.40 & - & 15.60 & 90.10 & - & 23.10 & 84.30 & - & 19.65 \\
DRDG (IJCAI'21)~\cite{liu2021dual} & 12.43 & 95.13 & - & 19.05 & 88.79 & - & 15.56 & 91.79 & - & 15.63 & 91.75 & - & 15.66 \\
ANRL (ACM MM'21)~\cite{liu2021adaptive} & 10.83 & 96.75 & - & 17.83 & 89.26 & - & 16.03 & 91.04 & - & 15.67 & 91.90 & - & 15.09 \\
SSDG-R (CVPR'20)~\cite{jia2020single} & 7.38 & 97.17 & - & 10.44 & 95.94 & - & 11.71 & 96.59 & - & 15.61 & 91.54 & - & 11.28 \\
SSAN-R (CVPR'22)~\cite{wang2022domain} & 6.67 & 98.75 & - & 10.00 & 96.67 & - & 8.88 & 96.79 & - & 13.72 & 96.63 & - & 9.81 \\
PatchNet (CVPR'22)~\cite{wang2022patchnet} & 7.10 & 98.46 & - & 11.33 & 94.58 & - & 13.40 & 95.67 & - & 11.82 & 95.07 & - & 10.91 \\
SA-FAS (CVPR'23)~\cite{sun2023rethinking} & 5.95 & 96.55 & - & 8.78 & 95.37 & - & 6.58 & 97.54 & - & 10.00 & 96.23 & - & 7.82 \\
IADG (CVPR'23)~\cite{zhou2023instance} & 5.41 & 98.19 & - & 8.70 & 96.44 & - & 10.62 & 94.50 & - & 8.86 & 97.14 & - & 8.39 \\
CA-MoEiT (IJCV'24)~\cite{liu2024moeit} & 2.88 & 98.76 & 81.43 & 7.89 & 97.70 & 61.33 & 6.18 & 98.94 & 80.00 & 9.72 & 96.22 & 49.44 & 6.67 \\
GAC-FAS (CVPR'24)~\cite{le2024gradient} & 5.00 & 97.56 & - & 8.20 & 95.16 & - & {4.29} & {98.87} & - & 8.60 & 97.16 & - & 6.52 \\
CFPL (CVPR'24)~\cite{liu2024cfpl} & 3.09 & 99.45 & \underline{94.28} & {2.56} & \underline{99.10} & \underline{66.33} & 5.43 & 98.41 & 85.29 & 3.33 & {99.05} & \underline{90.06} & 3.60 \\
S-CPTL (ACM MM'24)~\cite{guo2024style} & \underline{1.43} & 99.17 & - & \underline{0.89} & 99.00 & - & 6.86 & 98.63 & - & {4.12} & 99.02 & - & {3.33} \\
FGPL (ACM MM'24)~\cite{hu2024fine} & 2.86 & 98.12 & - & 3.89 & 98.19 & - & \underline{3.50} & \underline{99.54} & - & \textbf{1.77} & \underline{99.23} & - & 3.01 \\
DiffFAS-V (ECCV'24)~\cite{ge2024difffas} & 2.86 & 98.41 & - & 10.11 & 96.32 & - & 6.36 & 97.89 & - & 8.11 & 97.27 & - & 6.86 \\
BUDoPT (ECCV'24)~\cite{liubottom} & \textbf{0.95} & \underline{99.70} & 87.12 & 2.85 & 98.03 & 55.00 & 4.4 & 98.54 & \underline{86.46} & \underline{2.26} & 98.78 & 55.25 & \underline{2.62} \\
\midrule
MVP-FAS (Ours) & {1.71} & \textbf{99.83} & \textbf{98.33} & \textbf{0.75} & \textbf{99.89} & \textbf{99.29} & \textbf{1.99} & \textbf{99.59} & \textbf{95.77} & {2.82} & \textbf{99.49} & \textbf{90.85} & \textbf{1.82} \\
\bottomrule
\end{tabular}
}
  \vspace{-0.2cm}
    % \caption{The results (\%) of Protocol 1 in comparison with SOTA models across the MSU-MFSD (\textbf{M}), CASIA-FASD (\textbf{C}), Replay-Attack (\textbf{I}), and OULU-NPU (\textbf{O}) datasets. Bold numbers highlight the best performance. }
    \caption{The results (\%) of cross-domain FAS methods on the MSU-MFSD (\textbf{M}), CASIA-FASD (\textbf{C}), Idiap Replay Attack (\textbf{I}), and OULU-NPU (\textbf{O}) datasets in Protocol 1. Bold and underlined numbers indicate the best and second-best performances, respectively. The symbols $\uparrow$ and $\downarrow$ indicate that higher and lower values are preferred, respectively.}
    \label{tab:sota_mcio}
\end{table*}

\subsection{Total Loss}
The total loss of MVP-FAS consists of a classification loss $\mathcal{L}_{\text{cls}}$ and the MTPA loss $\mathcal{L}_{\text{MTPA}}$ for patch alignment. 
Following prior work~\cite{liu2024cfpl, srivatsan2023flip}, we define the classification loss $\mathcal{L}_{\text{cls}}$ as the CE loss between the ground-truth $y$ and the final prediction $\hat{y}$:
\begin{equation}
\label{eqn: classification_loss}
\mathcal{L}_{\text{cls}} = \frac{1}{B}\sum_{k=1}^{B} {CE}(y_{k}, \hat{y}_{k}). 
\end{equation}
Finally, the total loss is given by the equation below.
\begin{equation}
\label{eqn: total_loss}
\mathcal{L}_{\text{total}} = \mathcal{L}_{\text{cls}} + \mathcal{L}_{\text{MTPA}}.
\end{equation}

\section{Experiments}
\label{sec:experiments}

\begin{table*}[hbt!]
    \centering
    \setlength\tabcolsep{2.5pt}\resizebox{\linewidth}{!}{
    \begin{tabular}{lccccccccccccc}
    \toprule
    \multirow{3}{*}{Method} & \multicolumn{3}{c}{CS $\rightarrow$ W} & \multicolumn{3}{c}{SW $\rightarrow$ C} & \multicolumn{3}{c}{CW $\rightarrow$ S} & {avg.} \\ 
    \cmidrule{2-11} 
    & HTER$\downarrow$ & AUC$\uparrow$ & \makecell{TPR@FPR=1\%} $\uparrow$ & HTER $\downarrow$ & AUC $\uparrow$ & \makecell{TPR@FPR=1\%} $\uparrow$ & HTER $\downarrow$ & AUC $\uparrow$ & \makecell{TPR@FPR=1\%} $\uparrow$ & HTER $\downarrow$ \\ 
    \midrule
    ViT (ECCV'22)~\cite{huang2022adaptive} & 21.04 & 89.12 & 30.09 & 17.12 & 89.05 & \underline{22.71} & 17.16 & 90.25 & 30.23 & 18.44 \\
    %CFPL쪽 성능 CLIP부분은 빼도되지 않을까
    CLIP-V (ICML'21)~\cite{radford2021learning} & 20.00 & 87.72 & 16.44 & 17.67 & 89.67 & 20.70 & \textbf{8.32} & \textbf{97.23} & {57.28} & 15.33 \\
    CLIP (ICML'21)~\cite{radford2021learning} & 17.05 & 89.37 & 8.17 & 15.22 & \underline{91.99} & 17.08 & 9.34 & 96.62 & \textbf{60.75} & 13.87 \\
    CoOp (IJCV'22)~\cite{zhou2022learning} & 9.52 & 90.49 & 10.68 & {18.30} & {87.47} & {11.50} & 11.37 & 95.46 & 40.40 & 13.06 \\
    CFPL (CVPR'24)~\cite{liu2024cfpl} & {9.04} & \underline{96.48} & 25.84 & 14.83 & 90.36 & {8.33} & \underline{8.77} & \underline{96.83} & 53.34 & {10.88} \\
    FGPL (ACM MM'24)~\cite{hu2024fine} & 14.05 & 92.65 & \underline{33.33} & 19.00 & 88.53 & 13.33 & 11.00 & 94.72 & 34.00 & 14.68 \\
    S-CPTL (ACM MM'24)~\cite{guo2024style} & \underline{8.99} & 94.01 & - & \underline{12.78} & 91.64 & - & 9.48 & 95.83 & - & \underline{10.42} \\
    \midrule
    MVP-FAS (Ours)& \textbf{4.57} & \textbf{98.91} & \textbf{73.70} & \textbf{8.27} & \textbf{97.17} & \textbf{52.14} & 10.19 & {96.35} & \underline{60.46} & \textbf{7.68} \\
    % \midrule

    %     ViT* (ECCV'22)~\cite{huang2022adaptive} & 7.98 & 97.97 & 73.61 & 11.13 & 95.46 & 47.59 & 13.35 & 94.13 & 49.97 & 10.82 \\
    % % ViTAF*-5-shot~\cite{huang2022adaptive} & 2.91 & 99.71 & 92.65 & 6.00 & 98.55 & 78.56 & 11.10 & 95.03 & 60.12 & 6.83 \\
    % FLIP-MCL* (ICCV '23)~\cite{srivatsan2023flip} & 4.46 & 99.16 & 83.66 & 9.66 & 96.69 & 50.11 & 11.71 & 95.21 & 57.98 & 8.61 \\
    % S-CPTL (ACM MM'24)~\cite{guo2024style} & 4.42 & 99.30 & - & 9.59 & 96.73 & - & 10.97 & 97.40 & - & 8.33 \\
    % CFPL* (CVPR'24)~\cite{liu2024cfpl} & 4.40 & 99.11 & 85.23 & 8.13 & 96.70 & 62.41 & 8.50 & 97.00 & 55.66 & 7.01 \\
    % MSPT* (Ours) & 4.80 & 98.34 & 42.63 & \textbf{4.01} & \textbf{99.32} & \textbf{86.89} & 13.38 & 93.89 & 52.71 & 7.40 \\
    \bottomrule
    \end{tabular}
    }
    % \caption{The results (\%) of Protocol 2 in comparison with SOTA models across the CASIA-SURF (S), CASIA-SURF CeFA (C), and WMCA (W) datasets. Bold numbers indicate the best performance for each metric. }
  \vspace{-0.2cm}
    \caption{The results (\%) of cross-domain FAS methods on the CASIA-SURF (\textbf{S}), CASIA CeFA (\textbf{C}), and WMCA (\textbf{W}) datasets in Protocol~2.}

    \label{tab:sota_scw}
    \vspace*{-.4cm}
\end{table*}

\subsection{Experimental Setup}
\textbf{Datasets and Protocols. }Following prior works on cross-domain FAS~\cite{srivatsan2023flip,liu2024cfpl,liubottom,hu2024fine,guo2024style}, we use two distinct protocols to assess the generalization performance of our proposed MVP-FAS.
In Protocol 1, we evaluate our method using four widely recognized FAS benchmark datasets: MSU-MFSD (\textbf{M})~\cite{wen2015face}, CASIA-FASD (\textbf{C})~\cite{zhang2012face}, Idiap Replay Attack (\textbf{I})~\cite{chingovska2012effectiveness}, and OULU-NPU (\textbf{O})~\cite{boulkenafet2017oulu}.
In Protocol 2, we use three large-scale FAS benchmark datasets: CASIA-SURF (\textbf{S})~\cite{zhang2020casia}, CASIA CeFA (\textbf{C})~\cite{liu2021casia}, and WMCA~(\textbf{W})~\cite{george2019biometric}, which include a large number of subjects, extreme domain variations, and extensive samples.
Both protocols follow a leave-one-out scheme, which assesses FAS performance on a target dataset based on models trained using the remaining datasets for each protocol.

\noindent\textbf{Evaluation Metrics. }For a robust measurement of generalization performance, we employ three widely used FAS metrics: Half Total Error Rate (HTER), Area Under the Curve (AUC), and True Positive Rate (TPR) at a 1\% False Positive Rate (FPR) (TPR@FPR=1\%).

\noindent\textbf{Implementation Details. }We use a pre-trained CLIP~\cite{radford2021learning} model with ViT-B/16~\cite{dosovitskiy2020image}.
For the image encoder, input images are resized to $224\times224$, resulting in $L=196$ patch embeddings with a patch size of 16. 
For the text encoder following CoOp~\cite{zhou2022learning}, we set the length of the learnable prompt $[\mathbf{V}]$ to 16 and use the same $[\mathbf{V}]$ for all classes.
We empirically set the number of positive and negative views to $M=3$. 
All feature dimensions in MVP-FAS are set to 512. 
MVP-FAS is trained for 30 epochs in Protocol 1 and 300 epochs in Protocol 2, both with a batch size of 18. 
We use the Adam optimizer~\cite{kingma2014adam} with a weight decay of 0.001 and initial learning rates of 1e-6 for Protocol 1 and 1e-7 for Protocol 2.

\subsection{Cross-domain FAS Performance}
\noindent\textbf{Protocol 1.}
Table~\ref{tab:sota_mcio} shows the results of MVP-FAS and recent SOTA methods for Protocol 1 across the \textbf{M}, \textbf{C}, \textbf{I}, and \textbf{O} datasets. 
MVP-FAS achieves the best average HTER among all methods, including recent CLIP-based models~\cite{liu2024cfpl,hu2024fine,guo2024style,liubottom}. 
Notably, MVP-FAS surpasses the previous SOTA method, BUDoPT~\cite{liubottom}, with a substantial 0.8 percentage points (\%p) improvement in average HTER, as well as gains in AUC and TPR@FPR=1\%.
MVP-FAS also reduces HTER deviations across scenarios to 0.7\%, compared to 1.2\% for BUDOPT.
Moreover, MVP-FAS shows a significant increase in the average TPR@FPR=1\%, improving from 83.99\% in CFPL~\cite{liu2024cfpl} to 96.06\%. 

\noindent\textbf{Protocol 2.}
We present the results for Protocol 2 in Table~\ref{tab:sota_scw}.
Similar to Protocol 1, MVP-FAS achieves the SOTA performance across all metrics on average. 
MVP-FAS demonstrates a considerable improvement, reducing the average HTER by 2.74\%p compared to the previous SOTA method, S-CPTL~\cite{guo2024style}.
Specifically, MVP-FAS significantly reduces HTER by 4.42\%p and 4.51\%p in the target domains \textbf{W} and \textbf{C}, respectively, outperforming S-CPTL.

Overall, the experimental results of Protocol 1 and Protocol 2 indicate that MVP-FAS generalizes well to diverse and challenging environments, achieving consistently high performance across all metrics. 
Moreover, its strong performance in TPR@FPR=1\% on cross-domain datasets underscores its reliability in high-security scenarios and its suitability for real-world FR systems.

\subsection{Ablation Study}

\begin{table}[t]
\centering
\setlength\tabcolsep{6pt}\renewcommand{\arraystretch}{.95}\resizebox{.95\linewidth}{!}{

\begin{tabular}{ccccccc}
\toprule
Baseline & MVS & MTPA & HTER$\downarrow$ & AUC$\uparrow$ & \makecell{TPR@FPR=1\%}$\uparrow$ \\
\midrule
\checkmark & - & - & 8.59 & 95.99 & 60.27 \\
% \checkmark & - & - & - & - & - \\
\checkmark & \checkmark & - & 7.48 & 96.87 & 72.77 \\
\checkmark & - &  \checkmark & 4.13 & 99.05 & 73.09 \\
\checkmark & \checkmark & \checkmark & \textbf{1.82} & \textbf{99.70} & \textbf{96.06} \\
\bottomrule
\end{tabular}
}
\vspace{-0.2cm}
% \caption{An ablation of each component. Each result is the average on all scenarios in Protocol 1. Bold numbers indicate the best performance for each metric.}
\caption{Ablation study on each component of MVP-FAS,  
%on each component, 
with results (\%) averaged across all scenarios in Protocol 1.}
% \caption{Ablation study on each component.}
\label{tab:ablation_component}
\vspace*{-.5cm}
\end{table}
We conduct all ablation studies on the cross-domain scenarios in Protocol 1. 
To ensure a precise and reliable comparison, all studies except for the single-view text pair comparison (Sec.~\ref{sec:prompt_analysis}) are evaluated based on the average scores of HTER(\%), AUC(\%), and TPR(\%)@FPR=1\%.
\subsubsection{Effectiveness of Each Component}
\label{sec:effectiveness_of_each_component}
To evaluate the impact of MVS and MTPA components in our MVP-FAS, we establish the baseline model using only the global-aware patch embeddings and our classification head on CLIP.
Each component is gradually added to this baseline, and we evaluate their performance.
As shown in Table~\ref{tab:ablation_component}, including MVS in the baseline reduces HTER by 1.11\%p. This demonstrates that MVS improves the model's generalization ability by capturing multi-view features derived from diverse texts, even if patch embeddings are misaligned.
Meanwhile, applying only MTPA leads to a 4.46\%p improvement in the baseline HTER. 
This suggests that patch-level alignment is essential to exploit the local information of patch embeddings in CLIP-based FAS models.
Finally, the integration of MTPA and MVS results in significant performance improvements across all metrics, with 6.77\%p, 3.71\%p, and 35.79\%p in terms of HTER, AUC, TPR@FPR=1\%, respectively. 
These results indicate that the combination of multi-view features and aligned patch embeddings synergistically improves generalization.

\subsubsection{Prompt Analysis for Multi-View Approach} 
\label{sec:prompt_analysis}
We use five pairs of positive and negative texts as candidates for the multiple texts in MVP-FAS.
The detailed configuration of these pairs is provided in the first column of Table~\ref{tab:ablation_prompt}.
Initially, we evaluate each pair in the single-view setting of the MVS to confirm individual properties, such as domain bias.  
Table~\ref{tab:ablation_prompt} shows the overall average and scenario-specific results on HTER. 
Among the single-view pairs, `real face $\leftrightarrow$ spoof face' achieves the best average HTER performance of 3.75\%. 
Each text pair performs better in specific domains.
For example, although the average performance for `genuine face $\leftrightarrow$ fake face' is relatively low with 3.93\%, it individually achieves the best score of 3.84\% for \textbf{OCM}$\rightarrow$\textbf{I}. 
Similarly, the `bonafide face $\leftrightarrow$ attack face' achieves the highest score of 2.63\% for \textbf{OCI}$\rightarrow$\textbf{M}.
However, in terms of generalization, using only one text pair may hinder the model's capability for cross-domain FAS.

\begin{table}[t]
\centering
    \setlength\tabcolsep{2.5pt}\resizebox{\linewidth}{!}{
\begin{tabular}{lcccccc}
\toprule
Single-view(+face) & \multicolumn{1}{c}{OCI $\rightarrow$ M} & \multicolumn{1}{c}{OMI $\rightarrow$ C} & \multicolumn{1}{c}{OCM $\rightarrow$ I} & \multicolumn{1}{c}{ICM $\rightarrow$ O} & \multicolumn{1}{c}{avg.} \\ 
\cmidrule{2-6} 
(pos$\leftrightarrow$neg) & HTER(\%)$\downarrow$ & HTER(\%)$\downarrow$ & HTER(\%)$\downarrow$ & HTER(\%)$\downarrow$ & HTER(\%)$\downarrow$ \\ 
\midrule
real $\leftrightarrow$ spoof & 4.33 & \textbf{2.15} & 5.11 & 3.40 & \textbf{3.75} \\
bonafide  $\leftrightarrow$ attack & \textbf{2.63} & 2.91 & 5.07 & 4.43 & 3.76 \\
genuine $\leftrightarrow$ fake & 5.00 & 3.55 & \textbf{3.84} & \textbf{3.31} & 3.93 \\
true $\leftrightarrow$ false & 3.29 & 2.85 & 5.71 & 4.58 & 4.11 \\
verified $\leftrightarrow$ deceptive & 5.79 & 2.56 & 3.99 & 4.43 & 4.19 \\
\midrule
\textbf{Ours ($M=3$)} & \textbf{1.71} & \textbf{0.75} & \textbf{1.99} & \textbf{2.82} & \textbf{1.82} \\
\bottomrule
\end{tabular}
}
  \vspace{-0.2cm}
% \caption{Comparison on HTER performance across M, C, I, and O for each single-view in Protocol 1. All single-view contains the word `face' after it.}
% \caption{Comparison of HTER among single-view text pairs in Protocol 1. Each text is concisely 
%  presented with `face' omitted.% for conciseness.
%  }
 \caption{Comparison of HTER among single-view text pairs in Protocol 1. Each text is concisely 
 presented with `face' omitted.% for conciseness.
 }
\label{tab:ablation_prompt}
\vspace*{-.2cm}
\end{table}
\begin{table}[t]
\centering
\setlength\tabcolsep{10pt}\renewcommand{\arraystretch}{.9}
      \resizebox{.9\linewidth}{!}{
% \begin{table}[h!]
\centering
\begin{tabular}{lccc}
\toprule
{Number} & {HTER}$\downarrow$ & {AUC}$\uparrow$ & {TPR@FPR=1\%}$\uparrow$ \\
\midrule
$M=1$ & 3.75 & 99.09 & 86.94 \\
$M=2$ & 2.89 (-0.86) & 99.47 (+0.38) & 91.75 (+4.81) \\
\textbf{$M=3$} & \textbf{1.82 (-1.93)}  & \textbf{99.70 (+0.61)} & \textbf{96.06 (+9.12)} \\
$M=4$ & 1.95 (-1.80) & 99.66 (+0.57) & 93.87 (+6.93)\\
$M=5$ & 1.97 (-1.78) & 99.63 (+0.54) & 92.84 (+5.90) \\
\bottomrule
\end{tabular}}
  \vspace{-0.2cm}
% \caption{Ablation study on the number of multi-views. Each result is the average across all scenarios in Protocol 1. Bold numbers indicate the best performance for each metric.}
\caption{Ablation study on the number of multi-views, with each result (\%) indicating the average across all scenarios in Protocol 1.}
% \caption{Ablation study on the number of multi-views.}
\label{tab:ablation_multiview}
\vspace*{-.5cm}
\end{table}

\subsubsection{Effectiveness of Multi-View Number} 
\label{sec:multi_prompt_analysis}
To investigate the optimal number of multi-views, we construct models that progressively incorporate each text pair in the order of higher single-view HTER (Sec.~\ref{sec:prompt_analysis}).
As shown in Table~\ref{tab:ablation_multiview}, all multi-view settings outperform the single-view setting ($M=1$).
Notably, multi-view performance improves even when incorporating a text pair that initially exhibits lower performance in the single-view setting.
These experimental results indicate that the multi-view setting enhances generalization ability for cross-domain FAS. 
The best overall performance is observed at $M=3$. 
However, it slightly declines as more text pairs are added ($M\geq4$), likely due to increased noise. 
Nevertheless, the multi-view settings consistently surpass the single-view settings.

\subsubsection{Types of MTPA Anchor}
To determine the optimal anchor value for patch alignment across multiple texts, we compare three methods: a single text pair `real $\leftrightarrow$ spoof', which shows the best performance in Table~\ref{tab:ablation_prompt} as the anchor
(Single), aligning all multi-text individually as anchors (Individual), and utilizing the mean value of the multi-text as the anchor (Mean). 
As shown in Table~\ref{tab:ablation_mtpa}, the Individual approach performs worse than using a single text, although it uses the same number of multi-texts. 
These results suggest that treating each text individually as an anchor causes them to act as outliers in the text embedding space, leading to lower performance. 
Using the mean value of multi-texts as an anchor, MTPA reduces the influence of these outliers, achieving a more generalized text representation compared to using a single text, thereby resulting in superior performance.

\begin{table}[h]
\centering
\setlength\tabcolsep{10pt}\renewcommand{\arraystretch}{.7}
      \resizebox{.9\linewidth}{!}{
% \begin{table}[h!]
\centering
\begin{tabular}{lccc}
\toprule
{Type} & {HTER}$\downarrow$ & {AUC}$\uparrow$ & {TPR@FPR=1\%}$\uparrow$ \\
\midrule
% Single & 4.27 & 98.85 & 82.54\\
Single & 3.98 & 99.15 & 75.84\\
Individual & 4.58 & 98.69
& 60.00 \\
\textbf{Mean (Ours)} & \textbf{1.82} & \textbf{99.70} & \textbf{96.06} \\
\bottomrule
\end{tabular}}
  \vspace{-0.2cm}
% \caption{Ablation study on MTPA, with each result representing the average performance for Protocol 1. Single indicates that the real $\leftrightarrow$ spoof case, which performs best in single-view~\ref{tab:ablation_prompt}}
\caption{Ablation study %on MTPA
on the anchor of MTPA, with each result (\%) representing the average performance in %for
Protocol 1. %`Single' denotes `real $\leftrightarrow$ spoof' text pair, which shows the best performance in Table~\ref{tab:ablation_prompt}.
}
\label{tab:ablation_mtpa}
\vspace*{-.5cm}
\end{table}

\subsubsection{Effectiveness of Slot Attention in MVS}
\label{sec:effectiveness_of_MVS}
\noindent As shown in Table~\ref{tab:ablation_attention}, we perform an ablation study to evaluate the effectiveness of slot attention in MVS, comparing it against two alternative methods that also incorporate text and patch information.
The first variant uses a similarity-based approach, predicting outcomes by calculating the cosine similarity between the mean of global-aware patch embeddings and text embeddings, similar to CLIP.
The second variant employs cross-attention, with texts as queries and patches as keys and values.
Even though they utilize the same global-aware patch embedding, slot attention surpasses other methods across all metrics.
This suggests that slot attention effectively captures global-local patch information through multi-view representation.

\begin{table}[h]
\centering
\setlength\tabcolsep{10pt}\renewcommand{\arraystretch}{.7} %.9
      \resizebox{.9\linewidth}{!}{
% \begin{table}[h!]
\centering
\begin{tabular}{lccc}
\toprule
{Method} & {HTER}$\downarrow$ & {AUC}$\uparrow$ & {TPR@FPR=1\%}$\uparrow$ \\
\midrule
Similarity & 5.26 & 98.50 & 76.37 \\
% Cross Attention(Q-patch) & 20.65 & 87.60 & 18.96 \\
Cross-attention & 5.68 & 98.31 & 74.97 \\
\textbf{MVS (Ours)} & \textbf{1.82} & \textbf{99.70} & \textbf{96.06} \\
\bottomrule
\end{tabular}}
  \vspace{-0.2cm}
%\caption{An ablation study to evaluate the contribution of the MVS by replacing it with a conventional attention method. The average performance of the three metrics in Protocol 1}
% \caption{Ablation study to evaluate the contribution of MVS.}
\caption{Ablation study to evaluate the impact of MVS, with each result (\%) indicating the average across all scenarios in Protocol 1.}
\label{tab:ablation_attention}
% \vspace*{-.3cm}
\vspace*{-.6cm}
\end{table}

\subsubsection{Impact of Global-Aware Patch Embedding}
Table~\ref{tab:ablation_global} presents a performance comparison between global-aware patch embedding and original patch embedding in MVS.
We observe that global-aware patch embedding outperforms regular patch embedding, suggesting that incorporating a CLS token to provide global context effectively improves overall performance.
We believe that global-aware patch embedding enables a more effective allocation of text embeddings in CLIP-based MVP-FAS, because the original CLIP is pre-trained to align the CLS token with text.
\begin{table}[h]
\centering
\setlength\tabcolsep{5pt}\renewcommand{\arraystretch}{.9}
      \resizebox{.8\linewidth}{!}{
% \begin{table}[h!]
\centering
\begin{tabular}{lccc}
\toprule
{} & {HTER}$\downarrow$ & {AUC}$\uparrow$ & {TPR@FPR=1\%}$\uparrow$ \\
\midrule
Patch embedding & 5.39 & 97.92 & 75.78 \\
{GAPE} & \textbf{1.82} & \textbf{99.70} & \textbf{96.06} \\
\bottomrule
\end{tabular}}
  \vspace{-0.2cm}
\caption{Ablation study on global-aware patch embedding (GAPE), with each result (\%) representing the average across all scenarios in Protocol 1.}
\label{tab:ablation_global}
% \vspace*{-.4cm}
\vspace*{-.6cm}

\end{table}

\begin{figure}[ht]
  \centering
    \includegraphics[width=\linewidth]
  {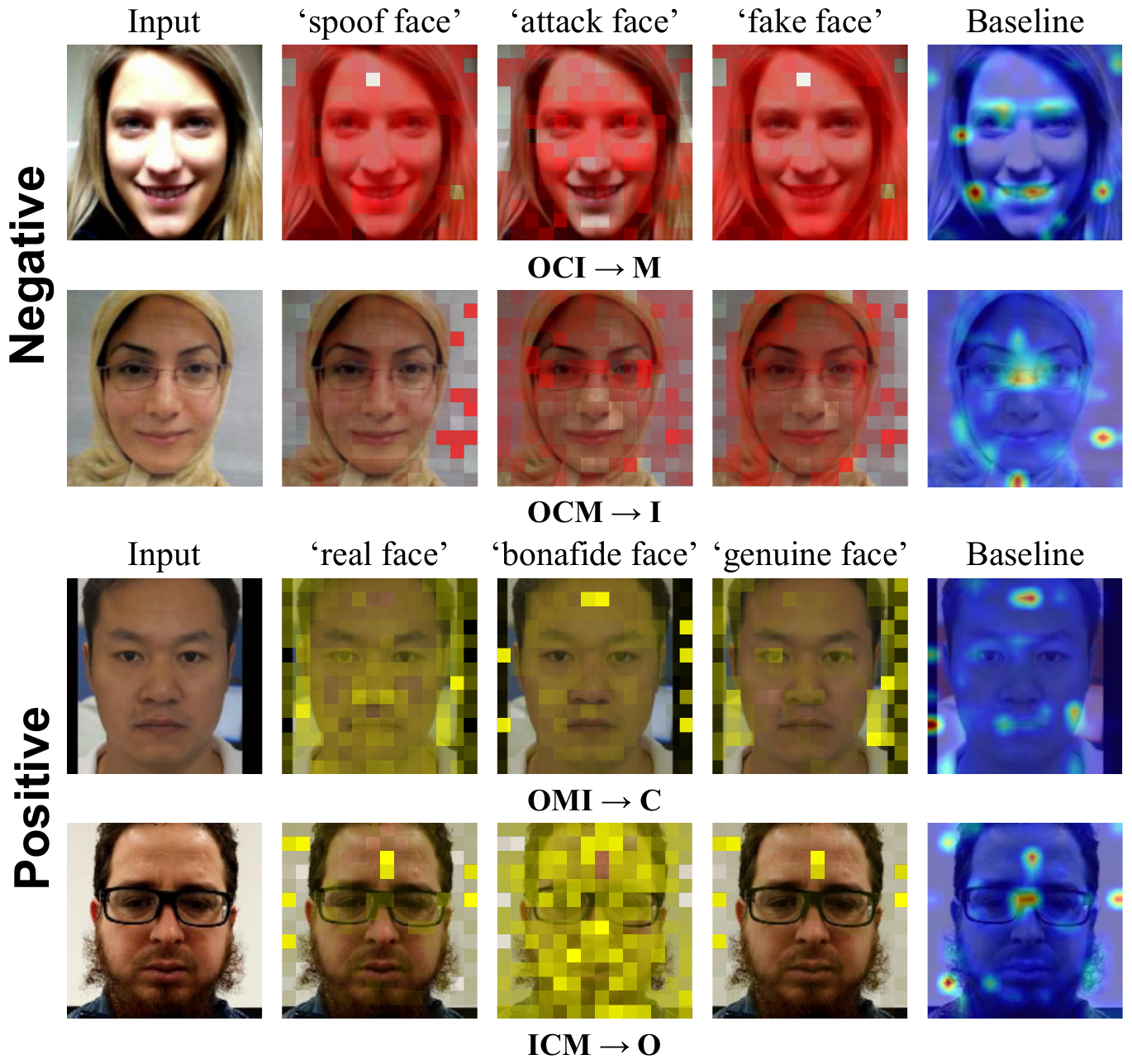}
  \vspace{-0.4cm}
  \caption{ 
  {\textbf{Visualization results for each view on real and spoof images across all sub-protocols in Protocol 1.} We visualize multi-view attention scores in the first stage of MVS. 
  Baseline (Sec.~\ref{sec:effectiveness_of_each_component}) indicates the attention map~\cite{chefer2021generic}, following the visualization method used in previous studies~\cite{srivatsan2023flip,liu2024cfpl,hu2024fine}. For better visibility, we visualize positive scores in yellow and negative scores in red.
  }}
  \label{fig:slot_visualization}
\vspace*{-.6cm} 
\end{figure}

\subsection{Visualization of Multi-Views}  
We visualize multi-view attention scores from MVS, providing more informative insights into predictions than the existing visualization method~\cite{chefer2021generic} used in prior works~\cite{srivatsan2023flip,liu2024cfpl,hu2024fine}. 
As shown in Fig.~\ref{fig:slot_visualization}, MVS's multi-view visualization illustrates the degree to which positive and negative texts are assigned across all patches, indicating the model's basis for determining spoofing or real. 
In negative samples ($\textbf{OCI} \rightarrow \textbf{M}$ and $\textbf{OCM} \rightarrow \textbf{I}$), the assigned texts primarily focus on the eye and mouth regions, some background areas, and facial edges that reveal depth differences. 
For positive samples ($\textbf{OMI} \rightarrow \textbf{C}$ and $\textbf{ICM} \rightarrow \textbf{O}$), MVS tends to focus on the overall texture or style of the image, as well as light reflections on features like the nose and forehead when identifying a real face.
By providing diverse positive and negative views, our multi-view visualization suggests a way to overcome the limitations of the existing attention map~\cite{chefer2021generic}, which lacks clarity in explaining the relevance of activated features.

\section{Conclusion}

We propose MVP-FAS, a novel framework that includes MVS and MTPA, both of which leverage paraphrased multi-text to enhance cross-domain generalization. 
To the best of our knowledge, MVP-FAS is the first VLM-based FAS method to incorporate slot attention.
Our MVP-FAS markedly outperforms existing VLM-based FAS methods across cross-domain datasets, achieving SOTA performance and providing interpretable visualizations.
These results demonstrate that capturing diverse fine-grained information and global context via MVS, while aligning patches through MTPA can significantly improve the model's generalization ability in FAS.
We believe that our method will benefit other computer vision tasks that require fine-grained recognition. 

\section{Acknowledgment}
This work was supported by the National Research Foundation of Korea (NRF) grant funded by the Korea government (MSIT) (No. 2023R1A2C200337911 and No. RS-2023-00220762) and the Institute of Information \& Communications Technology Planning \& Evaluation (IITP) grant funded by the Korea government (MSIT) (No.RS-2025-02303870, Software Technology for Efficient Multimodal Visual Information Processing in High-Speed Spatial Interactions) and (No.RS-2024-00439762, Developing Techniques for Analyzing and Assessing Vulnerabilities, and Tools for Confidentiality Evaluation in Generative AI Models).
{
    \small
    \bibliographystyle{ieeenat_fullname}
    \bibliography{main}
}

\end{document}